# Machine vision for vial positioning detection toward the safe automation of material synthesis

*Leslie Ching Ow Tiong[1†] Hyuk Jun Yoo,[1,2†] Na Yeon Kim,[1,3] Kwan-Young Lee,[2*] Sang Soo Han,[1*], Donghun Kim,[1*]*

[1]Computational Science Research Center, Korea Institute of Science and Technology, Seoul 02792, Republic of Korea

[2]Department of Chemical and Biological Engineering, Korea University, Seoul 02841, Republic of Korea

[3]Department of Chemistry, Korea University, Seoul 02841, Republic of Korea

†These authors contributed equally.

*Correspondence to: donghun@kist.re.kr (D.K.); sangsoo@korea.ac.kr (S.S.H.); kylee@korea.ac.kr (K.-Y.L.)

# Abstract

Although robot-based automation in chemistry laboratories can accelerate the material development process, surveillance-free environments may lead to dangerous accidents primarily due to machine control errors. Object detection techniques can play vital roles in addressing these safety issues; however, state-of-the-art detectors, including single-shot detector (SSD) models, suffer from insufficient accuracy in environments involving complex and noisy scenes. With the aim of improving safety in a surveillance-free laboratory, we report a novel deep learning (DL)-based object detector, namely, DenseSSD. For the foremost and frequent problem of detecting vial positions, DenseSSD achieved a mean average precision (mAP) over 95% based on a complex dataset involving both empty and solution-filled vials, greatly exceeding those of conventional detectors; such high precision is critical to minimizing failure-induced accidents. Additionally, DenseSSD was observed to be highly insensitive to the environmental changes, maintaining its high precision under the variations of solution colors or testing view angles. The robustness of DenseSSD would allow the utilized equipment settings to be more flexible. This work demonstrates that DenseSSD is useful for enhancing safety in an automated material synthesis environment, and it can be extended to various applications where high detection accuracy and speed are both needed.

# Main

Automated chemical synthesis based on robotics and artificial intelligence has facilitated the material development process[1,2]. Recent examples involved the synthesis of a wide range of materials, such as organic/polymer materials[3–9], quantum dots[10–13], and nanoparticles[14]. Although automation can substantially increase developmental efficiency, it is often accompanied by severe dangers in situations where corrosive or inflammable chemicals are handled without human surveillance. Any accidents would cause significant losses of life and property and result in some causalities in severe cases. To democratize automation in material synthesis laboratories, safety-related issues such as machine control errors need to be addressed.

Automated chemical synthesis for batch processes, especially wet chemical synthesis, inevitably involves frequent movements of chemical vessels such as flasks[15–18], beakers[19–21], and vials[22,23]. If these vessels are incorrectly placed, any subsequent actions, such as solution stirring, may lead to undesired dangers. In this regard, detecting the movements of chemical vessels is considered an important task for improving safety in automated chemical synthesis, and deep learning (DL)-based computer vision can play a vital role here. Computer vision has been widely used in object segmentation and detection[24–28] in industries such as autonomous vehicles[29,30], disease diagnosis[31], and rehabilitation[32]. However, object detectors for automated chemical synthesis have yet to be reported, and we aim to develop a high-performance detector that is suitable for identifying the error positions of chemical vessels for safety purposes in the present study.

Generally, object detection is described as a collection of related computer vision tasks that involve identifying objects in the given image. Currently, models such as You Only Look Once (YOLO) and the single-shot detector (SSD) are the most popular DL-based object detectors introduced by Redmon et al.[33] and Liu et al.[34], respectively. These models perform well regarding the speed of detection in real-life scenarios. However, one of the challenges is that neither YOLO nor the SSD can achieve sufficiently high detection accuracy in complex scenes with noisy background images. In addition, these models also rarely focus on complex objects independently due to the lack of aggregating and exploring the information between the network layers. These limitations of YOLO and the SSD make it difficult to directly implement them in surveillance-free chemistry laboratories where very high detection precisions are required for safety purposes; this calls for the development of an improved detection model.

To achieve better detection accuracy and to maintain a high detection speed in an automated chemical synthesis laboratory, we report a novel object detector, namely, a densely connected single-shot detector (DenseSSD) with a densely connected mechanism[35]. The advantage of this model is that we proposed a densely connected pyramidal layer in the model structure so that it can be more robust when learning feature representations, and it achieved improved object detection performance over that of YOLO and the SSD. Owing to these benefits, DenseSSD achieved 95.2% mean average precision[35] (mAP) on a complex dataset involving both empty and solution-filled vials, greatly exceeding the values produced by YOLO and the SSD by 53.4% and 18.9%, respectively. In addition, DenseSSD was found to be comparatively insensitive to the environmental changes, maintaining the highest precision under the variations of solution colors or testing view angles. Such robustness of DenseSSD allows the equipment settings to be more flexible in laboratory environments. Last, to fully realize the potential of DenseSSD, we designed the safety alert module to remotely and immediately notify researchers of possible dangers when any failures are detected by DenseSSD. DenseSSD can be effectively extended to other detection tasks where both high detection accuracy and speed are necessary.

## Results

**Workflow and model description.** The overall workflow of vial positioning detection based on machine vision is schematized in Fig. 1a. The hardware consists of a vial storage box, a robot arm, multiple vial holders on the stirrer, and a camera taking a bird's-eye view of all these pieces of equipment and their related actions. Images of these hardware systems are provided in Supplementary Fig. S1. The robot arm is programmed to repeatedly attempt to move the vials in the storage box right into the vial holders on the stirrer. In automated synthesis environments where researcher surveillance is absent, vial movement attempts may fail, possibly due to robot malfunctioning, programming errors, and any external environmental changes. Even if they may occur with a very low probability, such errors may cause dangerous situations in synthesis laboratory environments; any subsequent actions must be immediately halted. In other words, when the machine vision process detects a failure case in the vial movement actions, the system instantly stops and is designed to remotely send an alert to responsible researchers. Otherwise, the system continues to perform the following action of moving the remaining vials. Supplementary Video 1 demonstrates the overall workflow of our system including hardware movements, object detection steps, and alert modules.

We propose a novel object detection model, namely, DenseSSD. The original SSD model was introduced by Liu et al.[34] for object detection and recognition, and the concept of a densely connected network[36] is incorporated in DenseSSD to extract comparatively richer feature representations for achieving enhanced vial positioning detection performance. In this study, we demonstrate that DenseSSD outperforms the benchmark models, including the original SSD and YOLO models, for the task of vial positioning detection. The YOLO model was the first attempt at building a DL-based object detector, and it was proposed by Redmon et al.[33]. As illustrated in Supplementary Fig. S2, the YOLO model utilizes DarkNet[37] as a backbone structure by adding four convolutional (*conv*) layers to explore the features of the entire input image and predict each bounding box candidate. This means that the model globally explores the full image and all the objects in the image without performing the region proposal step. However, the limitation is that such a model lacks the ability to recognize irregularly shaped objects or groups of small objects due to its lack of exploring the global and local features of the image. To overcome this limitation, Liu et al.[34] proposed the SSD by using pyramid feature representation layers to explore and correlate the global and local features for efficiently detecting objects. Here, the global features describe the entire image with general information, such as shape information, and the local features describe the image patches with specific details, such as texture information. The SSD model was inspired by the Visual Geometry Group (VGG)[38] model, which was designed as the base model for extracting useful image features. Specifically, as shown in Supplementary Fig. S2, the SSD adds several feature layers with decreasing sizes; these layers are defined as pyramid representations of images at different scales. Such a pyramid structure performs better than YOLO in capturing global and local features from the different scales of the representation layers to target objects of various sizes. However, this model hardly focuses on complex objects independently due to the lack of aggregating and exploring the information between the subsequent layers.

Considering the limitations above, we introduce a novel densely connected mechanism in the pyramid hierarchy structure to explore and aggregate the correlations of relevant features among the subsequent layers in the SSD model, namely, DenseSSD. The network architecture of DenseSSD is illustrated in Fig. 1b. DenseSSD utilizes the design concept of a densely connected mechanism[36], in which all layers are densely connected to extract comparatively global and local feature representations; in contrast, in the original SSD and YOLO models (Supplementary Fig. S2), the features in each *conv* layer are used as inputs for the next layer without communication. DenseSSD contains two components: a mainstream network and a pyramidal feature hierarchy structure. The mainstream network consists of four dense block (DB) layers and four transition layers. The DB layers use different connectivity patterns by introducing direct connections from any layer to all subsequent layers, which improves the information flow between layers. Each layer has access to all the preceding feature maps (FMs) in its blocks and thus to the network's collective knowledge. Next, the densely connected mechanism is additionally deployed in the pyramidal feature hierarchy structure to extract the multiscale FMs from different layers. The structure consists of six feature block (FB) layers as pyramidal layers and five reduction layers. The FB layers are devised to aggregate the multiple features derived from different regions and progressively explore the unique global and local features at each pyramidal layer. The reduction layer consists of an average pooling layer (*avgpool*) and *conv* 1x1 layer to remain the depth of FB representations. Owing to the densely-connected FB layers, DenseSSD is expected to collect rich information while maintaining low feature complexity. The details of the network configurations are provided in Supplementary Table S1.

**Object detection performance.** We collected a total of 789 images in our dataset. Each image contained several vials, which were manually labeled as either success or failure cases, as described in Fig. 2a and Supplementary Fig. S3. Note that success cases referred to vials correctly located in their holder, whereas failure cases referred to vials in any undesired places, namely, those in *fall-out*, *lie-down*, *lean-in*, and *stand-on* situations. Data augmentations methods based on image processing, including random flipping, brightness, saturation, hue, and Gaussian filter, were applied only to the images in the training dataset to overcome the imbalanced environmental factors (Supplementary Fig. S4). As a result, for the detector performance evaluations, 8,764 vial cases were used for training, and 1,502 cases were used for testing. The details about the dataset construction process are provided in Supplementary Tables S2 and S3.

DL experiments based on the image datasets were carried out to compare the vial positioning detection performance between DenseSSD and other benchmark models (YOLO[33] and the SSD[34]). The AP was used to measure the accuracy of the object detectors[35]. In Fig. 2b, DenseSSD achieved APs of 99.9% and 90.5% for each success and failure class and an mAP of 95.2%, greatly exceeding the mAPs of the SSD (82.1%) and YOLO (61.4%). In particular, DenseSSD outperformed the original SSD by 16.9% for failure cases, and such high detection precision for failure cases is required to minimize failure-induced accidents. Supplementary Video 2 describes a specific example where a scene involving eight vials was used to test three models, including DenseSSD. We observed that DenseSSD performed perfectly, while some detections provided by YOLO and the SSD were misleading. An error in the SSD case was that two bounding boxes with different labels were made for a *stand-on* vial. In the YOLO case, many error types were found as follows: (1) some vials were missed in the detection results, (2) failure cases were incorrectly classified as success cases, and (3) the bounding boxes were poorly made (small IoU value). In addition to the detection accuracy, the computational and memory efficiency levels of these models are compared in Fig. 2b. DenseSSD was superior to the SSD and YOLO in terms of both space complexity (total parameters) and time complexity (FLOPS: floating-point operations per second). The numbers of total parameters and FLOPS were 7.9 M and 19.2 M, respectively, which were significantly smaller than those of the SSD (parameter size=23.9 M, FLOPS=22.5 M) and YOLO (parameter size=320.6 M, FLOPS=24.3 In DenseSSD, the complexity is substantially relieved by optimizing the parameters and simplifying the connectivity between layers because it is unnecessary to learn redundant FMs. These comparisons indicated that the object detection process is extremely fast and efficient in our DenseSSD model.

To analyze the stability of the models, precision-recall (PR) curves were obtained, as shown in Fig. 2c. The analysis was performed by measuring the area under the curve (AUC)[39]. For reference, stability is a crucial factor that increases safety by reducing the false alarm probability, which refers to incorrectly identifying a failure case as a success case. As seen in Fig. 2c, DenseSSD effectively maintained its performance when applied to real-time vial positioning detection by achieving 0.97 as its AUC value; however, the SSD and YOLO only achieved 0.75 and 0.49 as their AUC values, respectively. This means that DenseSSD maintained its AP without less degradation than other models and accurately predicted bounding boxes and detections. Through this analysis, we conclude that DenseSSD exhibited significantly higher stability in detecting failure cases than the SSD and thereby provides a reasonable false alarm probability.

Some detection examples are provided in Fig. 2d and Supplementary Fig. S5, where most of the failure cases were accurately detected by DenseSSD without false alarm issues, in contrast,

YOLO and the SSD did not detect several failure cases well. In particular, as can be seen in Fig. 2d and Supplementary Figs. S5a-S5c, we noticed that vials positioned between vial holders were difficult to capture with the SSD due to their transparency, but they were successfully detected by DenseSSD. This indicates that DenseSSD is advantageous in differentiating the transparent objects exhibiting low brightness from the background. In addition, we also find in Fig. 2d and Supplementary Fig. S5d, that vials positioned on vial holders (either *lean-in* or *stand-on* failure cases) were erroneously detected by SSD with two bounding boxes for each object; the phenomenon was fixed with DenseSSD. We speculate that these benefits were likely derived from densely connected mechanism and pyramidal feature hierarchy in DenseSSD, which could explore and correlate the global and local features for object detections regardless of low brightness and noisy background.

To further understand the reasons why DenseSSD outperformed the other methods, we visualize the FMs of the three models in Fig. 3 and Supplementary Fig. S6. The FMs were captured in the FBs of the DenseSSD model and the *conv* layers of the SSD and YOLO models (Fig. 1b and Supplementary Fig. S2), and they look substantially different. For DenseSSD, the FMs focus on the localized regions of vials, and in particular, FM2 of DenseSSD successfully distinguishes two failure cases from the remaining success cases. In contrast, the FMs of SSD are obviously less clear, and those of YOLO are irregular and noisy, which is consistent with the higher performance of our DenseSSD model. These comparisons indicate that DenseSSD built clearer and richer FM representations that enhanced the pyramidal feature hierarchy structure to differentiate between the multiple vial positions. This is likely because unlike the competing approaches, DenseSSD receives direct supervision and reuses the feature patterns from the previous layers in its FBs, which share collective knowledge to improve the detection performance of the overall method.

**Testing view angle sensitivity.** To understand the robustness of DenseSSD to camera angle variations, we modified the datasets by adding images taken from different angles. The four angles of investigation included 30°, 45°, 60°, and 90°, as illustrated in Fig. 4a. The modified dataset included 2,377 original images across different angles. The data augmentation processes, such as image flipping, brightness, hue, and Gaussian blurring augmentation, were applied only to the images in the training dataset (Supplementary Fig. S4). As a result, for the detector performance evaluations, 32,715 vial cases were used for training, and 3,648 cases were used for testing. More details about the dataset construction process are provided in Supplementary Tables S2 and S3.

Transfer learning was performed by using the pretrained weights from the dataset with only 45° angles. In Fig. 4b, we observe that DenseSSD achieved the best mAPs on the testing datasets corresponding to different angles compared to those of the SSD and YOLO: 88.5% (30°), 94.8% (45°), 93.8% (60°), and 84.9% (90°). The mAP values of the SSD and YOLO are provided in Supplementary Table S4. Considerable mAP reductions at 30° and 90° was observed for all models. DenseSSD maintained its high mAP (over 93%) between 45° and 60°, indicating that its performance is highly insensitive to the testing view angle, and thus the model is robust to angle variations. Unlike that of DenseSSD, the mAPs of the other methods, including YOLO and the SSD, hit their highest values at only 45° and noticeably dropped at both 30° and 60°; these techniques are comparatively sensitive to the provided view angles. This test revealed that DenseSSD is robust to environmental changes. This insensitivity of DenseSSD is beneficial in a chemical laboratory because it allows the utilized equipment settings to be more flexible. The mAP values of YOLO, SSD, and DenseSSD are provided in Supplementary Table S5. The detection precisions achieved by DenseSSD with angle variations can further be understood with specific examples in Figs. 4b and 4c, involving two failure vial positioning cases with the *lean-in* and *stand-on* statuses. At 30°, two vials appeared to overlap, making the differentiation of these two objects highly difficult; as a result, the *lean-in* vial was incorrectly detected as the success case. Similarly at 90°, the *lean-in* vial was detected as success, likely due to the very similar appearances of these two types of vials from the top view. Testing view angles between 45° and 60° could prevent these limitations and led to enhanced mAPs exceeding 93.8%.

**Application involving solution-filled vial datasets.** Since vials are often filled with solutions in chemistry laboratories[40–43], the study needed to be expanded to datasets involving solution-filled vials. The positioning failures of solution-filled vials would be more dangerous than those of empty vials because the liquids in these vials may pour out and jeopardize neighboring equipment. Therefore, we performed object detection experiments by constructing more complex datasets involving both empty and solution-filled vials. Fig. 5a describes the composition of the training dataset used for the detection experiments. While the empty vial dataset was the same as the one used in the experiments in Fig. 2, we additionally collected 359 images for the dataset of solution-filled vials. Here, the colors of the filled solutions were randomly designed to make the detectors more robust to color changes. The vials in each image were labeled as either success or failure cases. The data augmentation procedures were also applied to the images in the training dataset, as described in Fig. 5b and Supplementary Fig. S7. As a result, a total of 17,174 vial cases were used for detector training, and 2,282 cases were used for testing. More details about the dataset construction process are provided in Supplementary Tables S2 and S3.

We first measured the precisions achieved when the models were trained only with the empty vial dataset, as shown in the upper panel of Fig. 5c. DenseSSD and the SSD achieved mAPs of only 81.2% and 75.5%, respectively, indicating that the empty vial dataset is not sufficient on its own for detecting the positions of solution-filled vials. In particular, DenseSSD only achieved a 67.5% AP for the failure cases, which is far from the satisfactory level. Next, we measured the precisions achieved when the models were trained with the full datasets containing both empty and filled vials. Very interestingly, the precision enhancements were dramatic for DenseSSD, whereas they were limited for the SSD. The mAP value of DenseSSD reached 95.2%, greatly exceeding the 76.3% of the SSD. Additionally, the AP of DenseSSD for the failure cases was improved to 90.8%, which was significantly higher than the 63.0% of the SSD, and such high precision is critical to minimizing failure-induced accidents. The mAP values of the SSD and YOLO are provided in Supplementary Table S6. The detection results obtained in the example scenes reveal the specific incorrect detection cases. For example, some solution-filled vials with the *stand-on* status were missed in the SSD detections. Additionally, multiple bounding boxes were formed near the vial with the *fall-out* status, causing misleading detection results. Overall, DenseSSD was demonstrated to be the most effective vial positioning detection method for both empty and solution-filled vials, greatly outperforming the existing SSD and YOLO models.

## Discussion

We have confirmed thus far that our DenseSSD model performs excellently in terms of vial positioning detection. Last, to fully realize its potential in surveillance-free chemistry laboratory environments, we additionally introduced a safety alert module that was designed to remotely notify researchers of possible dangers immediately after failure cases are detected by DenseSSD. Supplementary Fig. S8 shows the scheme of the alert module. Upon the detection of failure cases, any hardware system operations are immediately halted, and the alert module remotely sends the scene image and related text such as the event time and problematic vial's information to the user's messenger based on TCP/IP network communications. Such an alert module is an indispensable component of an automated laboratory environment because it is critical to minimizing failure-related losses. Supplementary Fig. S9 describes our alert module, which was tested with several popular messengers around the globe, including Facebook Messenger and Telegram.

In summary, we developed a DL-based object detector, namely, DenseSSD, which was demonstrated to accurately detecting chemical vials' positions in chemistry laboratory environments. Recently, the automation of chemical synthesis has garnered much attention mainly due to its potential to significantly increase material development efficiency; however, safety issues have rarely been addressed today. In the present study, DenseSSD significantly outperformed the previous detectors of YOLO and SSD, exhibiting detection precisions over 95% for the complex datasets involving both empty and solution-filled vials. The enhanced precision will no doubt contribute to minimizing the possible losses of life and property in surveillance-free laboratory environments. Our DenseSSD model is a general object detector and is not limited to chemistry-relevant datasets; thus, it will be useful for other detection tasks, such as self-driving vehicles, medical imaging, and remote sensors, where high detection accuracy and speed are both needed.

## Methods

**Hardware.** The hardware systems consisted of a vial storage box, a robot arm, and eight vial holders on a stirrer, as described in Supplementary Fig. S1. A camera was also set up for monitoring the equipment and their interactions. The vial storage box was prepared to supply vials, and the robot arm was programmed to relocate the vials in the box to the holders on the stirrer.

**Object detection models.** Regarding the configuration of DenseSSD, the model contains five FB layers and five reduction layers, as shown in Fig. 1b. Let an FB be a FM block with $l$ layers of $H$ that are composed of DBs and rectified linear unit layers:

$$\text{FB} = H_l([b_0, b_{l-1}]),$$

where $b_0$ and $b_{l-1}$ represent the transition and DB layers, respectively. The operator $[\cdot]$ is defined as a concatenation operator. Then, a reduction layer is implemented in the early stage of the FB and performs an average pooling operation and a 1×1 *conv* layer to aggregate the FM representations with the same dimensionality. Here, the 1×1 *conv* states that the filter size of the *conv* layer is 1×1. All the configurations are listed in Supplementary Table S1. To achieve the training objective, we define a total loss[34] ($L_{\text{total}}$) function as a weighted sum of the localization ($L_{\text{loc}}$) loss and the confidence loss ($L_{\text{conf}}$) as follows:

$$L_{\text{total}}(x, c, l, g) = \frac{1}{N}\left(L_{\text{conf}}(x, c) + \alpha L_{\text{loc}}(x, l, g)\right),$$

where $x, c, l,$ and $g$ are defined as the input image, multiclass confidence scores, predicted box, and ground-truth box, respectively. $N$ is defined as the number of matched boxes, and $\alpha$ is the weight for the localization loss. In this work, $\alpha$ is set as 0.5. Specifically, $L_{\text{loc}}$ is presented as follows:

$$L_{\text{loc}}(x, l, g) = \sum_{i \in Pos}^{N} \sum_m x_{ij}^k \, smooth_{L1}\left(l_i^m - g_j^m\right),$$

where $smooth_{L1}(\cdot)$ is defined as the smooth L1 loss[44] to calculate the localization loss between the predicted boxes ($l$) of $m$ and the ground-truth ($g$) boxes of $m$. Here, $m \in \{cx, cy, w, h\}$, where $cx$ and $cy$ are defined as the center of the default bounding box; $w$ and $h$ are defined as the width and height of the bounding box, respectively. In addition, $x_{ij}^k$ is an indicator for matching the $i^{\text{th}}$ predicted box to the $j^{\text{th}}$ ground-truth box of category $k$. Next, $L_{\text{conf}}$ is the softmax loss over multiple classes of confidence ($c$):

$$L_{\text{conf}}(x, c) = -\sum_{i \in Pos}^{N} x_{ij}^k \log\left(\hat{c}_i^p\right) - \sum_{i \in Neg}^{N} \log\left(c_i^0\right),$$

$$\hat{c}_i^p = \frac{\exp(c_i^p)}{\sum_p \exp(c_i^p)},$$

where $\hat{c}_i^p$ is known as the softmax loss function; $p$ is defined as a predicted box with a specific class, and 0 refers to a predicted box as a negative sample or background information.

During the training process, we applied the stochastic gradient descent (SGD) optimizer[45] with a learning rate of $1.0 \times 10^{-3}$, a weight decay of $1.0 \times 10^{-8}$ and a momentum of 0.9. In our experiments, the batch size was set to 64, and the training procedure was carried out for 500 epochs. Training was conducted on our dataset and was performed by following the protocols that were mentioned in the "Model structure and performance comparison" section. We also randomly divided the images in the training set for cross-validation purposes by taking 20% of the images as the validation set and the remaining images as the training set. Note that the model was trained using an Nvidia Tesla V100 GPU.

For YOLO and the SSD, we utilized all the existing models that were provided by their respective authors. We therefore made our best effort to modify the networks from the existing models by following YOLO[33] and the SSD[34] to perform fine-tuning by conducting training with our dataset, as mentioned earlier. We also applied the SGD optimizer with a learning rate of $1.0 \times 10^{-3}$, a weight decay of $1.0 \times 10^{-8}$ and a momentum of 0.9. In these experiments, the batch size was set to 64, and the training process was carried out for 500 epochs. The models were trained using an Nvidia Tesla V100 GPU.

**Evaluation metrics for object detection.** AP is a popular metric for measuring the accuracy of object detection models. The calculation of AP involves only one class. The metric can be calculated using the following equation:

$$\text{AP} = \sum_{i=1}^{n-1}(r_{i+1} - r_i)pre(r_{i+1}),$$

where $r_1, r_2, \ldots, r_n$ are the recall levels at which the precision (*pre*) is first interpolated. Next, the mAP is the average AP across all classes, which can be defined as follows:

$$\text{mAP} = \frac{1}{M}\sum_{i=1}^{M} \text{AP}_i,$$

where $M$ is defined as the number of classes.

## Data Availability

Several examples of the real-time scenes are available in the GitHub repository (https://github.com/KIST-CSRC/DenseSSD/tree/main/dataset/test_sample). The full dataset is only available by an email request.

## Code Availability

The code for the pre-trained model of DenseSSD is available in the GitHub repository (https://github.com/KIST-CSRC/DenseSSD). All the codes are written in Python 3.7 and the architecture of DenseSSD is implemented using PyTorch 1.7.

## Acknowledgements

This work was supported by the National Research Foundation of Korea funded by the Ministry of Science and ICT (NRF-2021M3A7C2089739 and NRF-2022M3H4A7046278).

## Author contributions

D.K. and S.S.H conceived the idea and supervised the project. L.C.O.T developed DenseSSD. L.C.O.T, H.J.Y., and N.Y.K. performed data collections, augmentations, labeling, and object detection experiments with DenseSSD, SSD, and YOLO. H.J.Y. developed the safety alert module. All authors contributed to results analysis and manuscript writing.

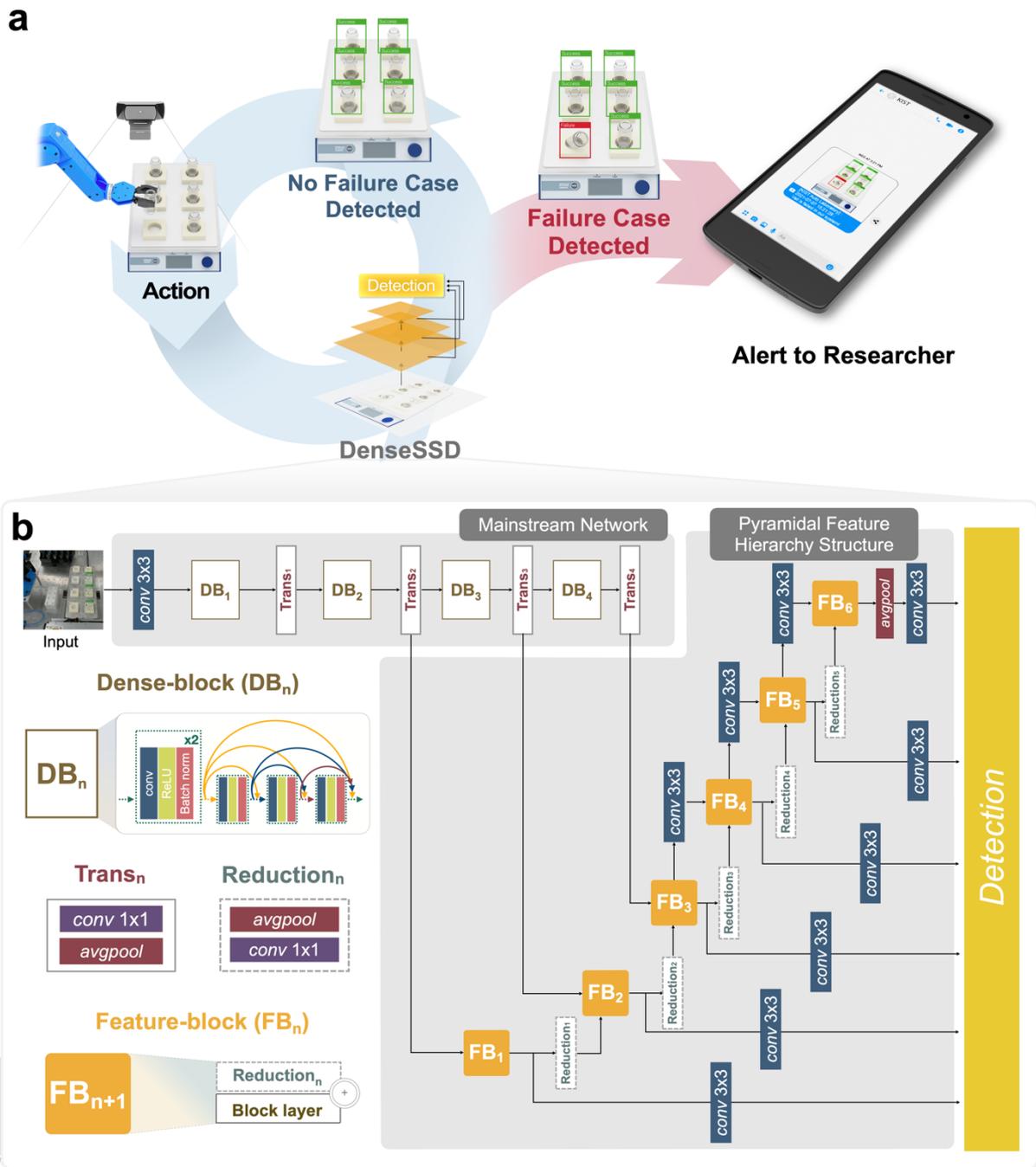

**Fig. 1 Workflow and model structure. a** Workflow of the vial positioning detection system based on DenseSSD. An action is defined as the movement of the robot arm to relocate each vial to its holder. Green and red boxes represent the predicted bounding boxes for the success and failure cases, respectively. When a failure vial positioning case is detected by DenseSSD, an alert is remotely and immediately sent to the responsible researcher. **b** The network architectures of DenseSSD, which largely consists of mainstream structure and pyramidal feature hierarchy structure. The *conv* 3×3 and *conv* 1×1 refer to convolution layers with filter sizes 3×3 and 1×1, respectively. The *avgpool* layer means an average pooling layer with its size 2×2. Each transition layer is composed of *conv* 1×1 and *avgpool*, while each reduction layer has the reversed order composition of *avgpool* and *conv* 1×1.

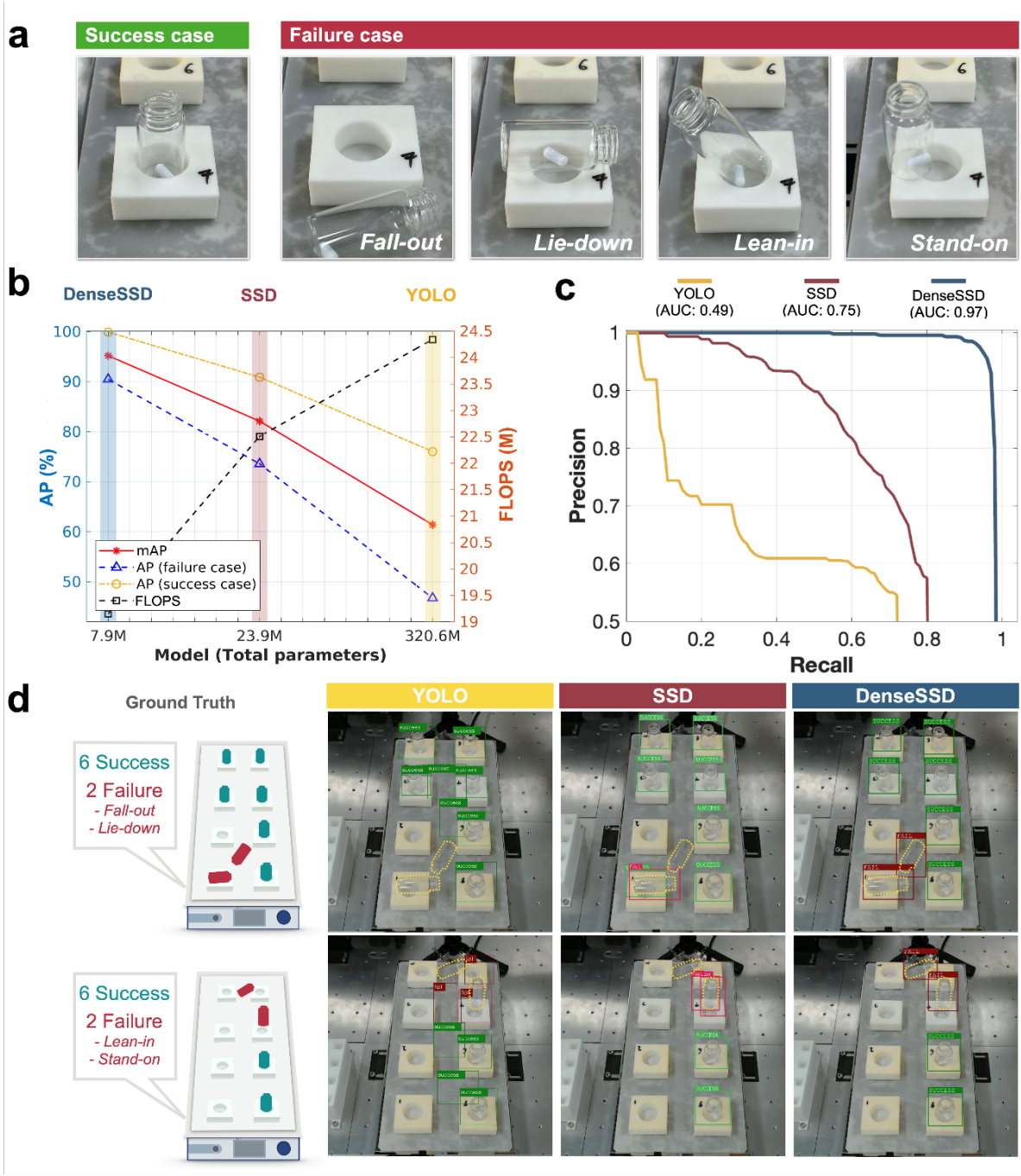

**Fig. 2 Vial positioning detection performance. a** Definitions and types of success and failure cases. Failure vial positioning involves four statuses of *fall-out*, *lie-down*, *lean-in*, and *stand-on*. **b** Detection performance of the DenseSSD, SSD, and YOLO models. AP, FLOPS, and total number of parameters of three models were compared. For AP evaluation, the threshold of 0.5 for the intersection over union (IoU) was used. **c** PR curves of the DenseSSD, SSD, and YOLO models. **d** Exemplary detection results of YOLO, SSD, and DenseSSD, achieved for two scenes. One scene contains eight vials including six success cases and two failure cases with the *fall-out* and *lie-down* statuses, while the other one contains four vials including two success cases and two failure cases with the *lean-in* and *stand-on* statuses. The failure vial cases are highlighted in yellow dotted lines for clarity.

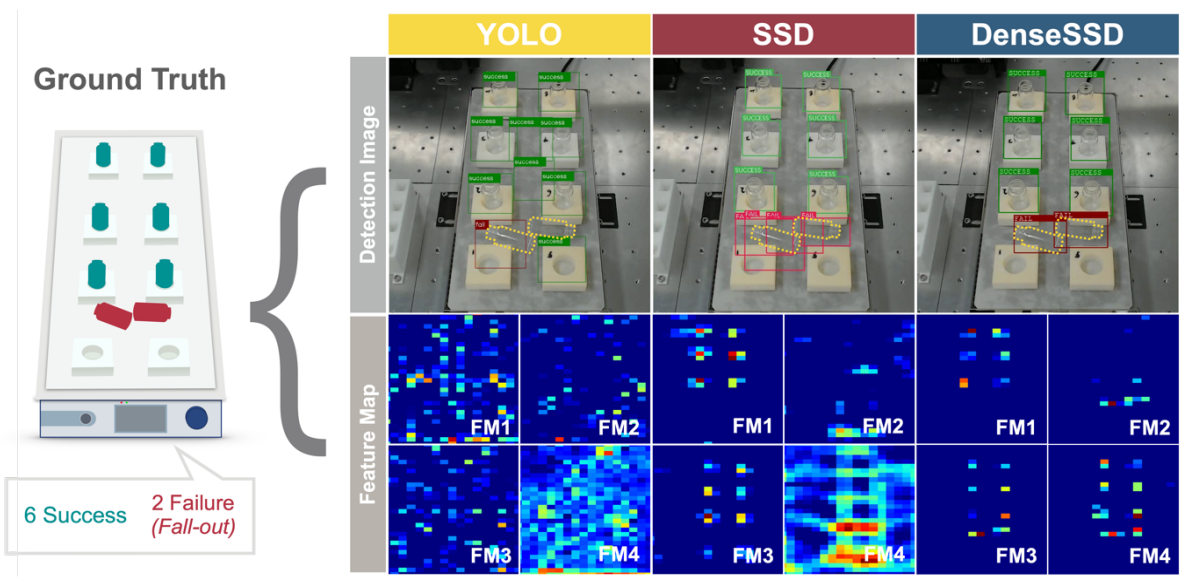

**Fig. 3 Visualization of the FM representations.** The ground-truth scene contains eight vials including six success cases and two failure cases with the *fall-out* status. The failure vial cases are highlighted in yellow dotted lines for clarity. The detection results and four FMs yielded by YOLO, the SSD, and DenseSSD are compared.

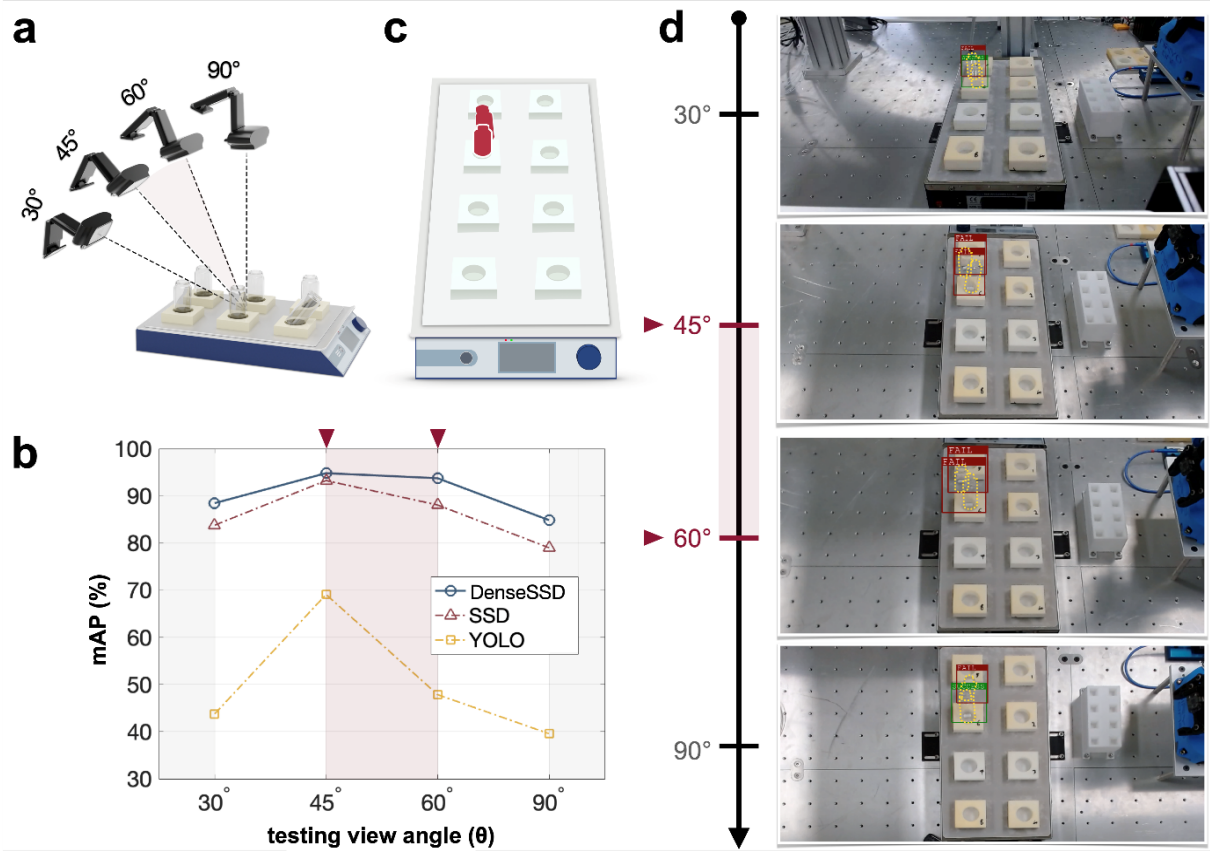

**Fig. 4. Testing view angle sensitivity**. **a** A scheme illustrating four camera angles (30°, 45°, 60°, and 90°). The origin for angle measurement is the center position of the stirrer. **b** Vial positioning detection performance (mAP) as a function of the testing view angle. **c** A scheme illustrating a ground-truth image involving two failure cases, one each with the *lean-in* and *stand-on* statuses. **d** The scenes and their corresponding detection results with DenseSSD for each tested camera angle. The failure vial cases are highlighted in yellow dotted lines for clarity. At 45° and 60°, both failure cases were correctly detected. At 30° and 90°, the vial in the *lean-in* status was incorrectly predicted as the success case.

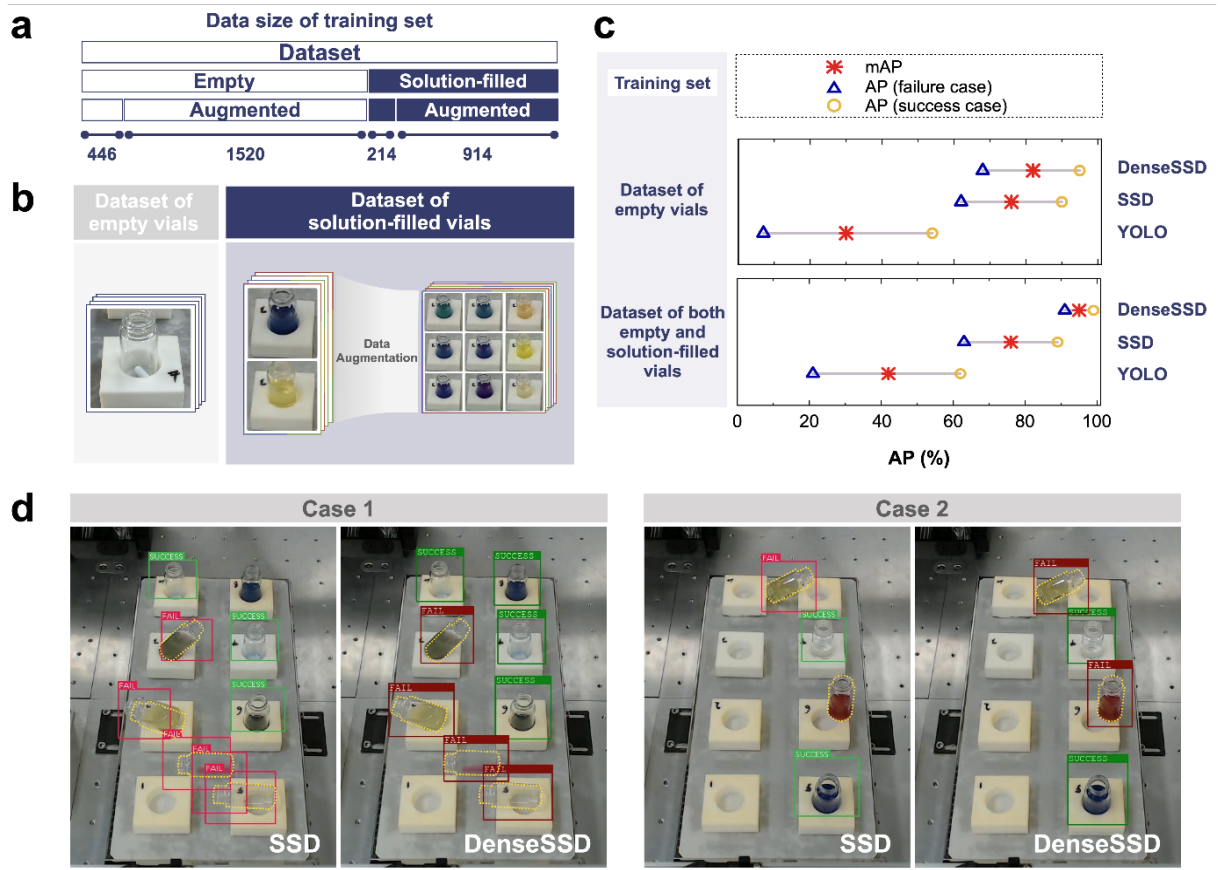

**Fig. 5 Application of DenseSSD to solution-filled vial datasets. a** Compositions of the training datasets used for the detection experiments. The number refers to the number of images in the training dataset. **b** The appearances of empty vials and solution-filled vials. Image processing techniques based on hue and saturations were applied to the images involving solution-filled vials for data augmentations. **c** Detection performance (AP) comparisons between YOLO, the SSD, and DenseSSD under different training set construction procedures: one that conducts training with only the empty vial dataset and another that conducts training with the dataset containing both empty and filled vials. **d** Exemplary detection results with SSD and DenseSSD, achieved for two scenes involving multiple solution-filled vials. One scene (Case 1) contains eight vials including four success cases and four failure cases with the *fall-out* and *lean-in* statuses, while the other one (Case 2) contains four vials including two success cases and two failure cases with the *stand-on* and *lean-in* statuses. The failure vial cases are highlighted in yellow dotted lines for clarity.

# Supplementary Information for

# Machine vision for vial positioning detections toward the safe automation of material synthesis

*Leslie Ching Ow Tiong[1†] Hyuk Jun Yoo,[1,2†] Nayeon Kim,[1,3] Kwan-Young Lee,[2*] Sang Soo Han,[1*], Donghun Kim,[1*]*

[1]Computational Science Research Center, Korea Institute of Science and Technology, Seoul 02792, Republic of Korea

[2]Department of Chemical and Biological Engineering, Korea University, Seoul 02841, Republic of Korea

[3]Department of Chemistry, Korea University, Seoul 02841, Republic of Korea

†These authors contributed equally.

*Correspondence to: donghun@kist.re.kr (D.K.); sangsoo@korea.ac.kr (S.S.H.); kylee@korea.ac.kr (K.-Y.L.)

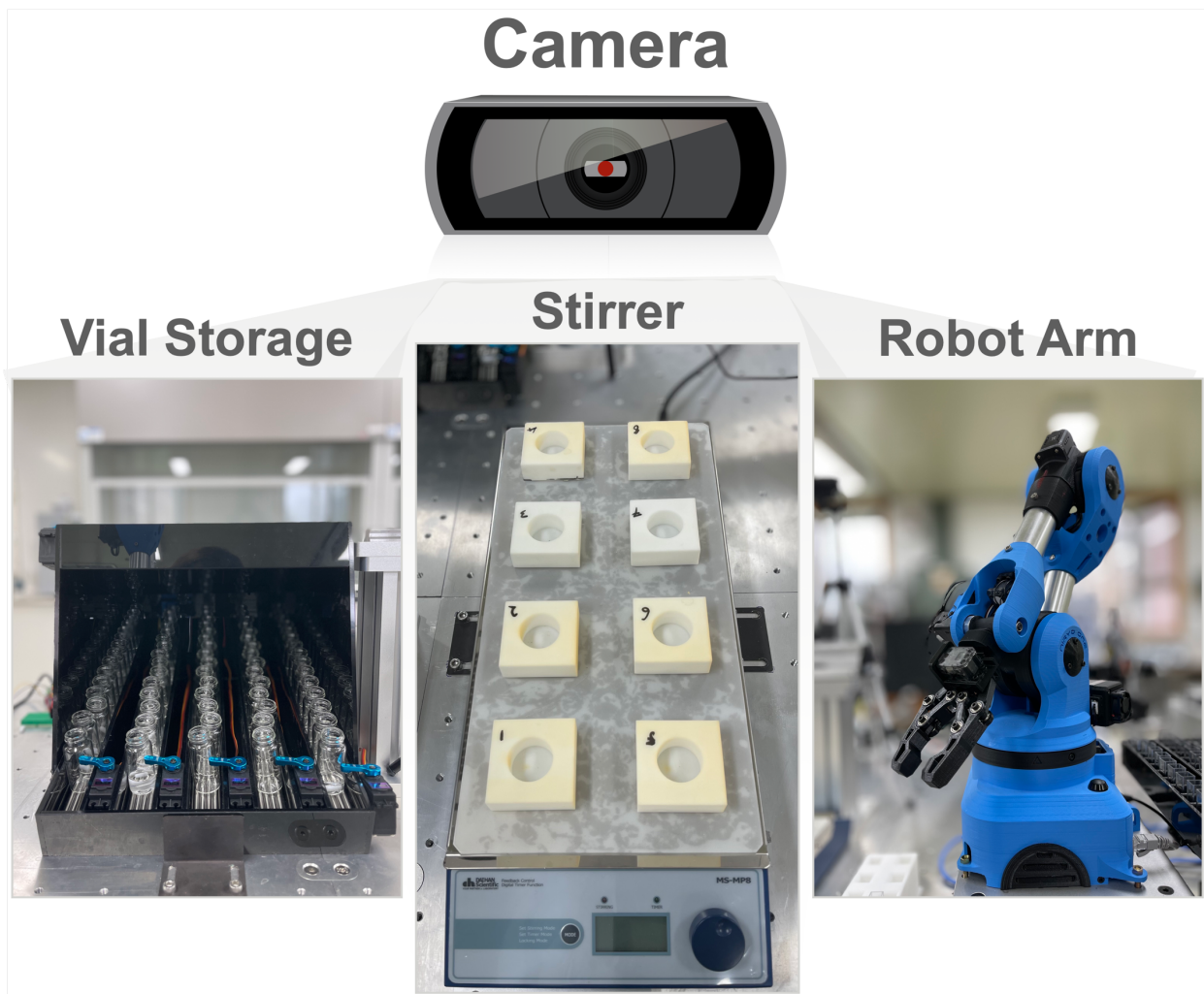

**Supplementary Figure S1. Schematic representation of hardware information** (web camera: logitech C920, vial storage, stirrer: DaiHan Scientific, robot arm: Niryo).

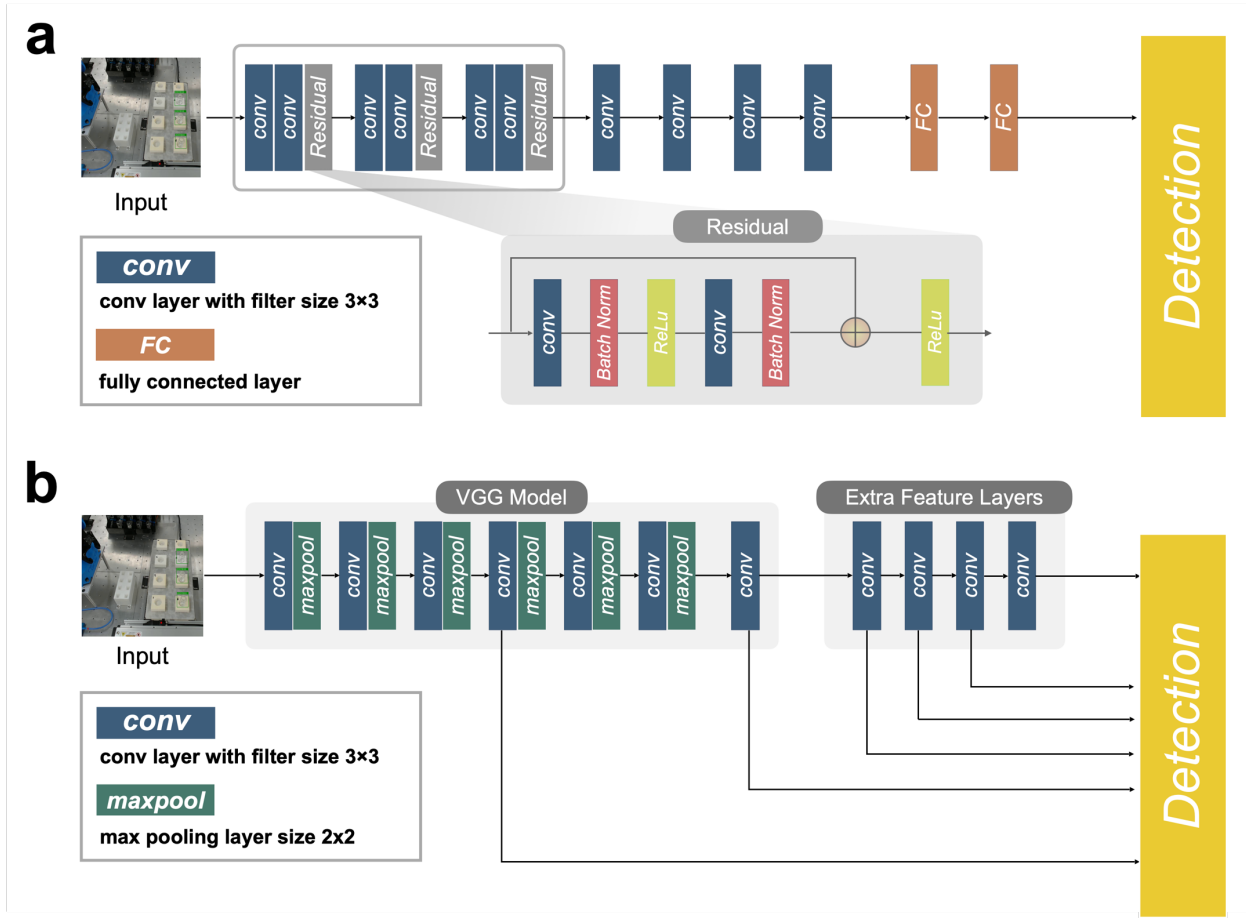

**Supplementary Figure S2. Architecture of benchmark models**; **a** YOLO; **b** SSD.

**Supplementary Table S1. The layers' information in DenseSSD**

| Main-stream Network | Layer Configurations | Pyramidal Feature Hierarchy Structure | Layer Configurations |
|---|---|---|---|
| Input | 300×300×3 | $FB_1$ | $\begin{bmatrix} Trans_2 \\ 3 \times 3\ conv \end{bmatrix}$ |
| $conv$ | 64@300 × 300; stride 2 | $Reduction_1$ | $2 \times 2\ avgpool; 1 \times 1\ conv;$ |
| $DB_1$ | $\begin{bmatrix} 1 \times 1\ conv \\ 3 \times 3\ conv \end{bmatrix} \times 6$ | $FB_2$ | $\begin{bmatrix} Reduction_1(FB_1), Tran_3 \\ 3 \times 3\ conv \end{bmatrix}$ |
| $Trans_1$ | $1 \times 1\ conv; 2 \times 2\ avgpool;$ stride 2 | $Reduction_2$ | $2 \times 2\ avgpool; 1 \times 1\ conv;$ |
| $DB_2$ | $\begin{bmatrix} 1 \times 1\ conv \\ 3 \times 3\ conv \end{bmatrix} \times 8$ | $FB_3$ | $\begin{bmatrix} Reduction_2(FB_2), Tran_4 \\ 3 \times 3\ conv \end{bmatrix}$ |
| $Trans_2$ | $1 \times 1\ conv; 2 \times 2\ avgpool;$ stride 2 | $conv_4$ | 128@3 × 3; stride 2 |
| $DB_3$ | $\begin{bmatrix} 1 \times 1\ conv \\ 3 \times 3\ conv \end{bmatrix} \times 16$ | $Reduction_3$ | $2 \times 2\ avgpool; 1 \times 1\ conv;$ |
| $Trans_3$ | $1 \times 1\ conv; 2 \times 2\ avgpool;$ stride 2 | $FB_4$ | $\begin{bmatrix} Reduction_3(FB_3), conv_4 \\ 3 \times 3\ conv \end{bmatrix}$ |
| $DB_4$ | $\begin{bmatrix} 1 \times 1\ conv \\ 3 \times 3\ conv \end{bmatrix} \times 6;$ | $Reduction_4$ | $2 \times 2\ avgpool; 1 \times 1\ conv;$ |
| $Trans_4$ | $1 \times 1\ conv; 2 \times 2\ avgpool;$ stride 2 | $conv_5$ | 128@3 × 3; stride 2 |
| | | $FB_5$ | $\begin{bmatrix} Reduction_4(FB_4), conv_5 \\ 3 \times 3\ conv \end{bmatrix}$ |
| | | $Reduction_5$ | $2 \times 2\ avgpool; 1 \times 1\ conv;$ |
| | | $conv_6$ | 128@3 × 3; stride 2 |
| | | $FB_6$ | $\begin{bmatrix} Reduction_4(FB_5), conv_6 \\ 3 \times 3\ conv \end{bmatrix}$ |
| | | Detection layer | $1 \times 1 \times 6$ |

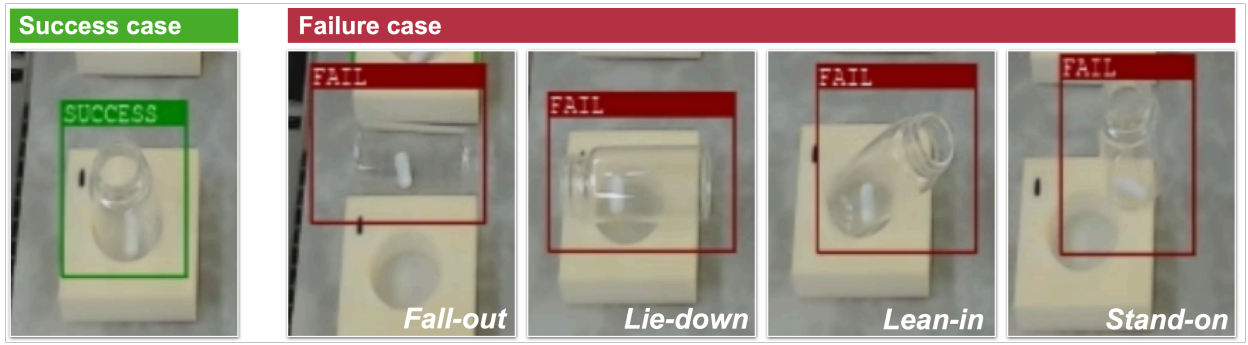

**Supplementary Figure S3.** The definition of success and failure cases and their detections with DenseSSD.

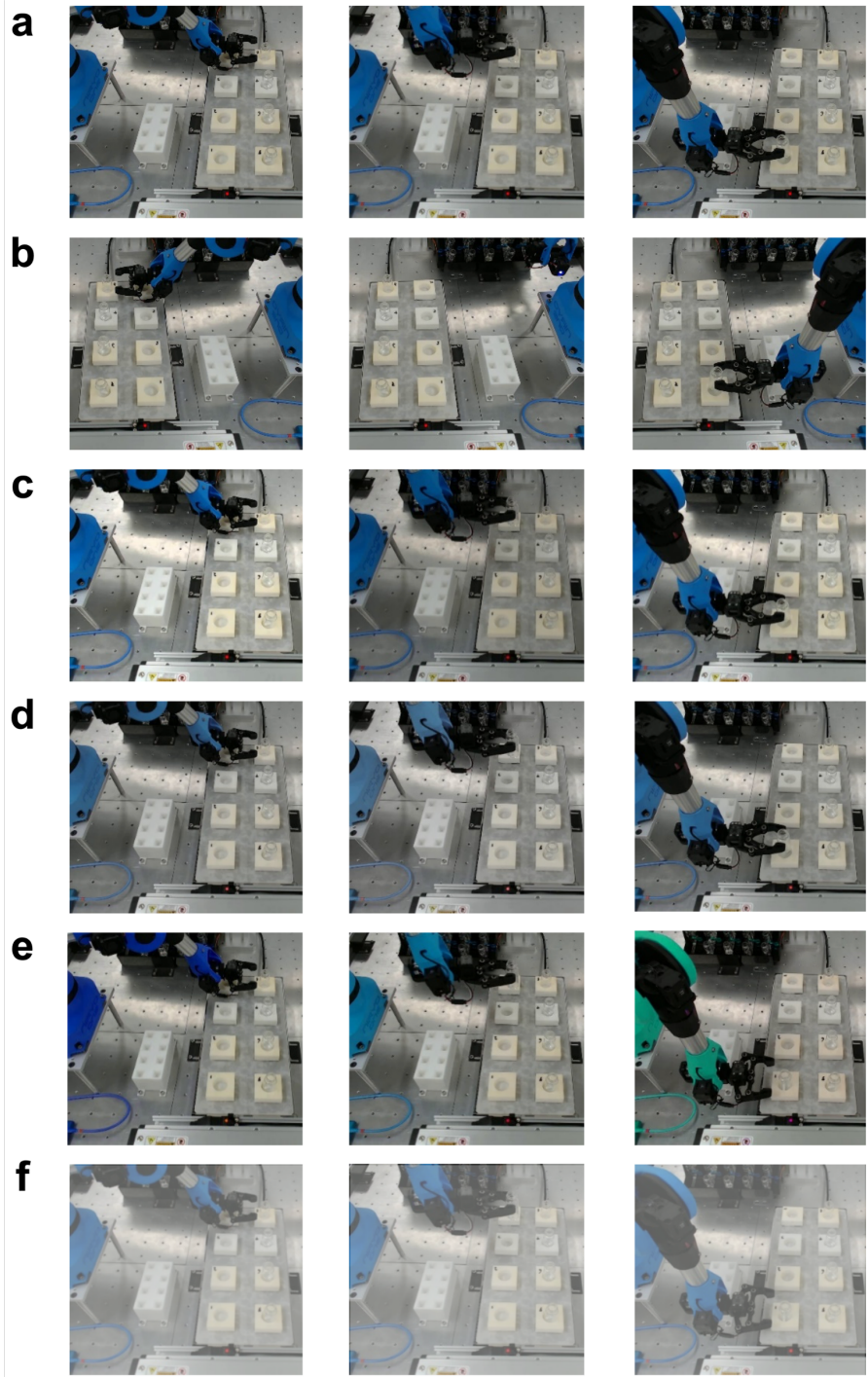

**Supplementary Figure S4. Real image of examples about data augmentation methods:** (a) Original, (b) Random flipping, (c) Brightness, (d) Saturation, (e) Hue, (f) Gaussian blurring.

**Supplementary Table S2. Dataset information (number of images)**

| Type | | Number of images | | |
|---|---|---|---|---|
| | *Augmentation* | *Training set* | *Testing set* | *Total* |
| Empty vials (45º) | No | 446 | 343 | 789 |
| | Yes | 1,966 | 343 | 2,309 |
| Solution-filled vials (45º) | No | 214 | 145 | 359 |
| | Yes | 1,128 | 145 | 1,273 |
| Empty vials (30º) | No | 392 | 179 | 571 |
| | Yes | 1,940 | 179 | 2,119 |
| Empty vials (60º) | No | 364 | 137 | 501 |
| | Yes | 1,794 | 137 | 1,931 |
| Empty vials (90º) | No | 358 | 158 | 516 |
| | Yes | 1,677 | 158 | 1,835 |

**Supplementary Table S3. Dataset information (number of vial cases)**

| Type | Augmentation | # of cases in training set | | # of cases in testing set | | Total | |
|---|---|---|---|---|---|---|---|
| | | *Success* | *Failure* | *Success* | *Failure* | *Success* | *Failure* |
| Empty vials (45°) | No | 1274 | 581 | 1099 | 403 | 2373 | 984 |
| | Yes | 6492 | 2272 | 1099 | 403 | 7591 | 2675 |
| Solution-filled vials (45°) | No | 918 | 780 | 437 | 343 | 1355 | 1123 |
| | Yes | 4542 | 3868 | 437 | 343 | 4979 | 4211 |
| Empty vials (30°) | No | 1252 | 458 | 324 | 277 | 1576 | 735 |
| | Yes | 6204 | 2249 | 324 | 277 | 6528 | 2526 |
| Empty vials (60°) | No | 1187 | 503 | 564 | 200 | 1751 | 703 |
| | Yes | 5723 | 2311 | 564 | 200 | 6287 | 2511 |
| Empty vials (90°) | No | 1015 | 433 | 452 | 379 | 1467 | 812 |
| | Yes | 5299 | 2165 | 452 | 379 | 5751 | 2544 |

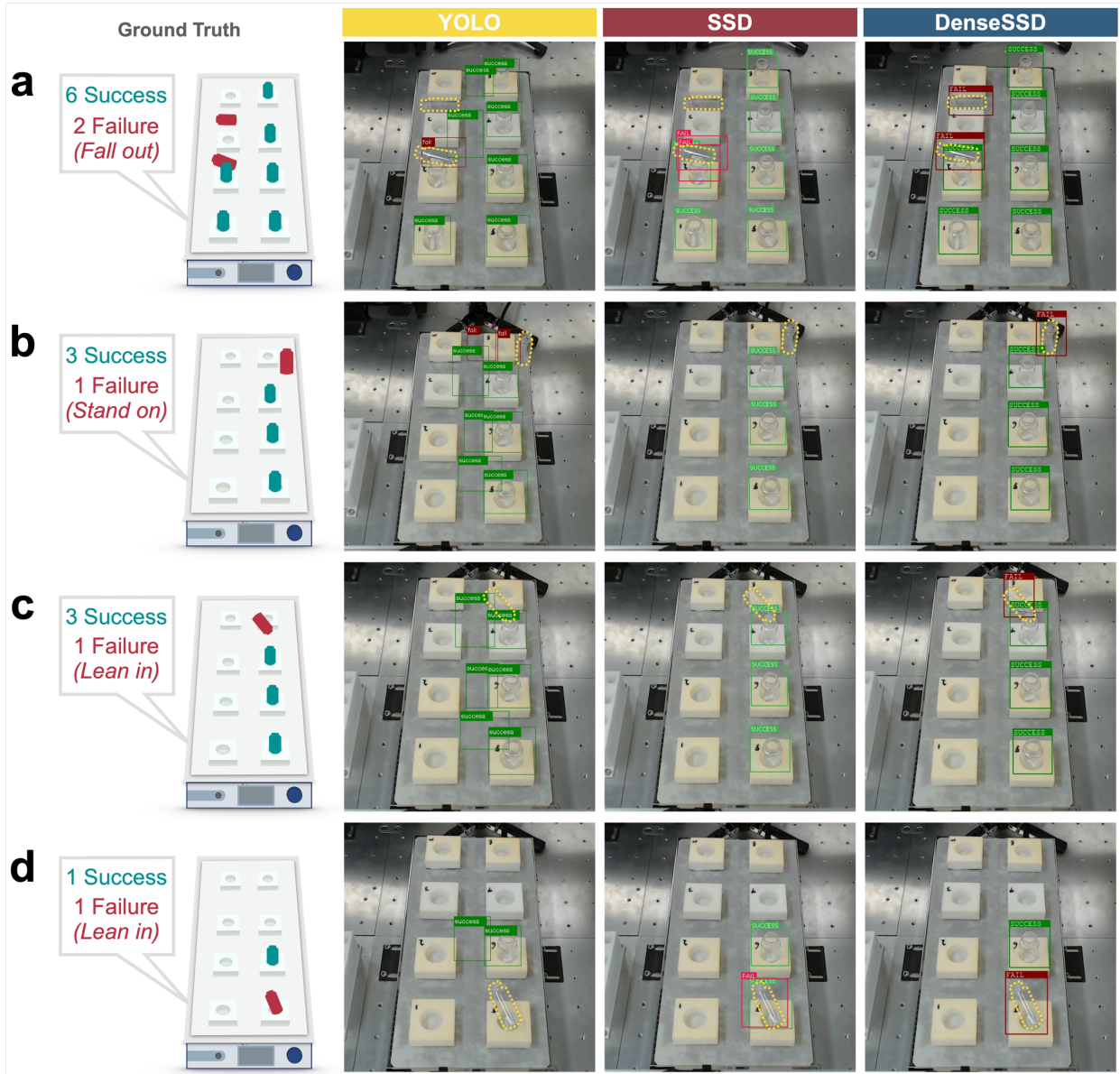

**Supplementary Figure S5. Demonstration of detection results with YOLO, SSD, and DenseSSD.** Yellow dotted circle refers to the failure cases are not detected by the benchmark models.

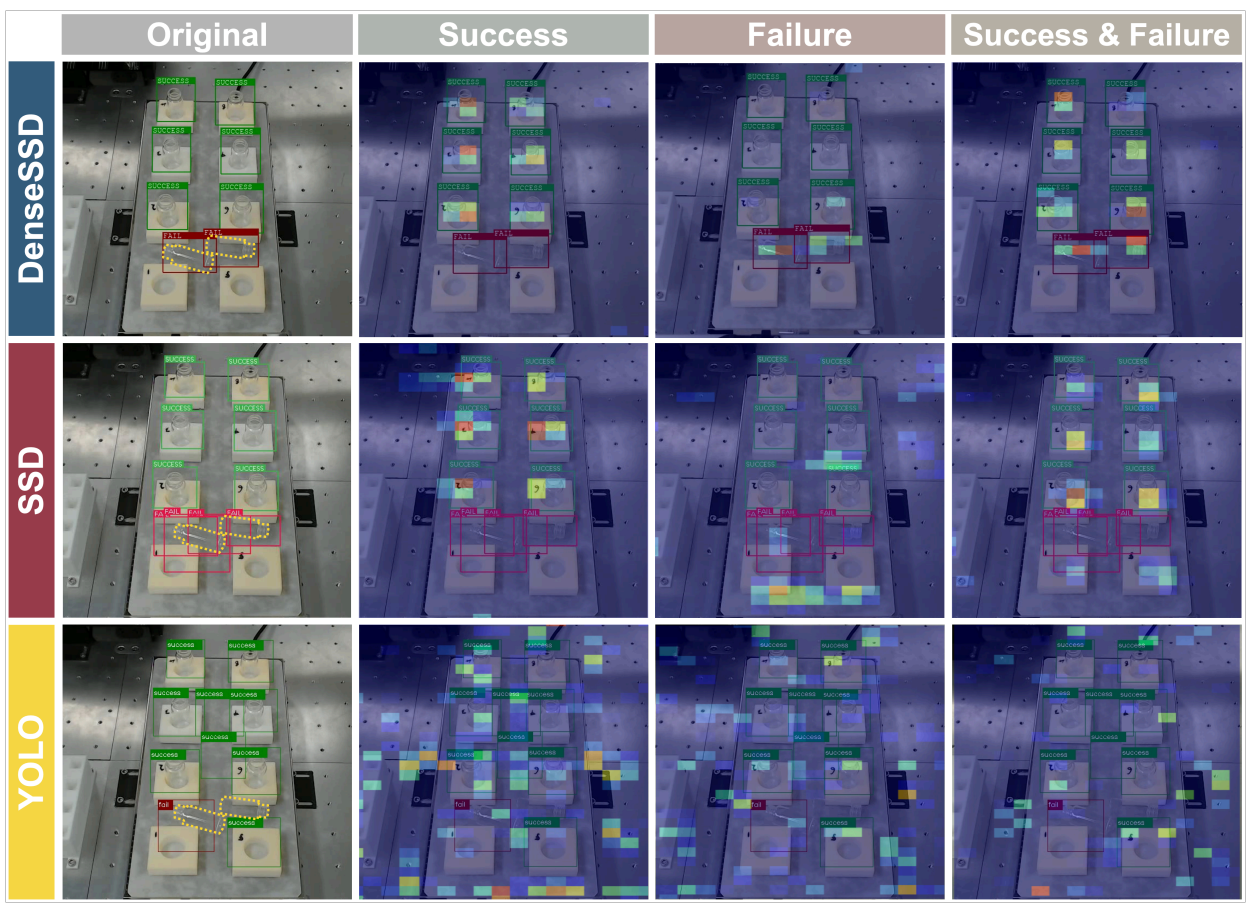

**Supplementary Figure S6. Visualization of feature map representations.** The original scene contains eight vials involving six success cases and two failure cases in the *fall out* status. Detection results and feature maps for YOLO, SSD, and DenseSSD are compared.

**Supplementary Table S4. Performance comparisons on vial positioning detection with camera angle 45º.** The highest precisions are written in bold.

| Model | AP (%) | | mAP (%) | FLOPS (M) | Parameter Size (M) |
|---|---|---|---|---|---|
| | *Success* | *Failure* | | | |
| YOLO | 76.0 | 46.8 | 61.4 | 24.3 | 320.6 |
| SSD | 90.9 | 73.6 | 82.1 | 22.5 | 23.9 |
| **DenseSSD** | **99.9** | **90.5** | **95.2** | **19.2** | **7.9** |

**Supplementary Table S5. Performance comparisons on vial positioning detection with different degrees of camera angles.** The highest mAPs are written in bold.

| Model | mAP (%) | | | |
| --- | --- | --- | --- | --- |
| | *30°* | *45°* | *60°* | *90°* |
| YOLO | 43.8 | 69.1 | 47.9 | 39.6 |
| SSD | 83.8 | 93.3 | 88.1 | 79.0 |
| **DenseSSD** | 88.5 | **94.8** | **93.8** | 84.9 |

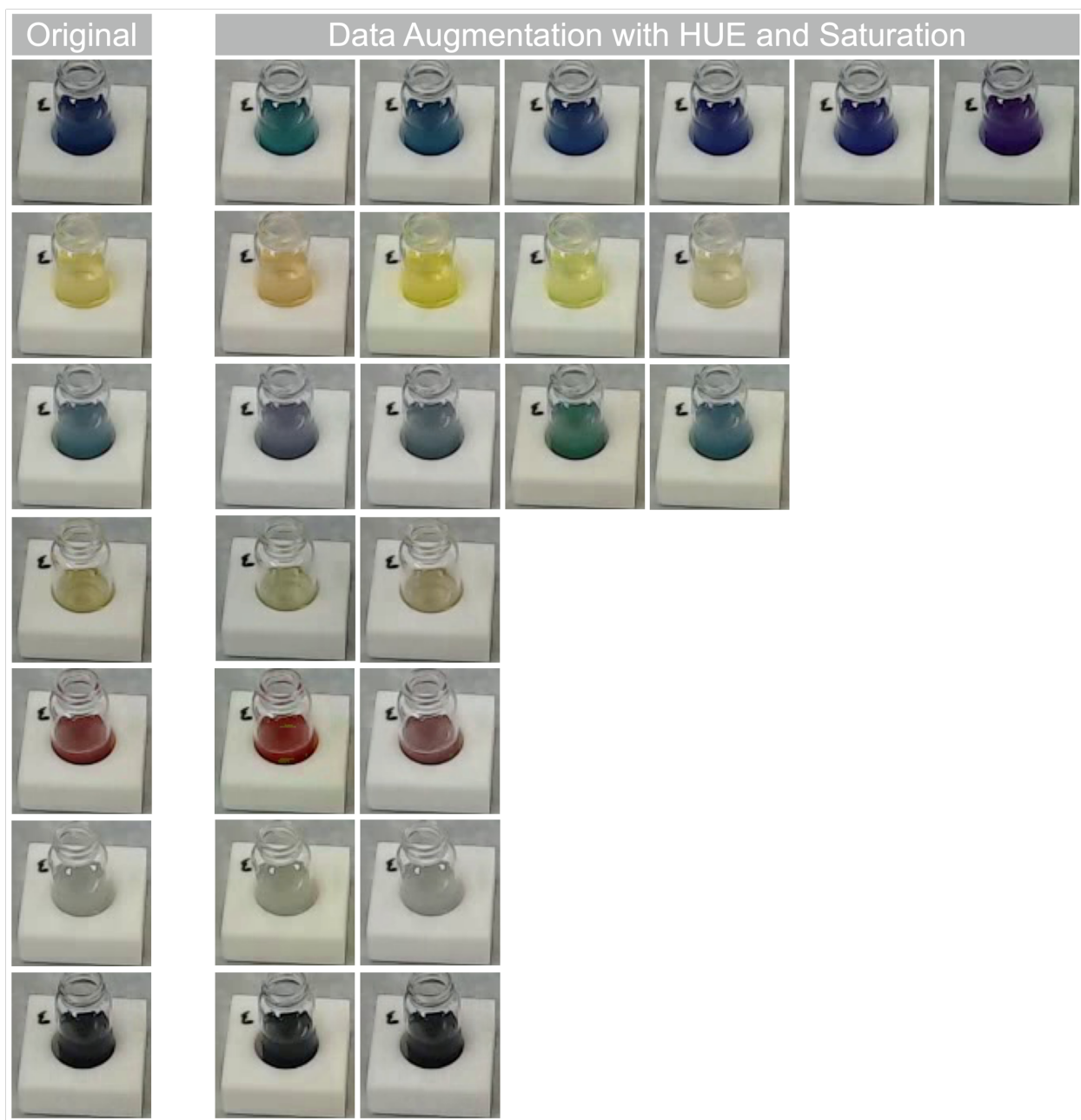

**Supplementary Figure S7.** Examples of data augmentation results (hue and saturation) for solution-filled vials.

**Supplementary Table S6. Performance comparisons on vial positioning detection with empty and solution-filled vials.** The highest precisions are written in bold.

| Model | Training set constructions | AP (%) Success | AP (%) Failure | mAP (%) |
|---|---|---|---|---|
| YOLO | Empty | 53.4 | 6.5 | 30.0 |
| | Empty + Solution-filled | 62.2 | 21.3 | 41.8 |
| SSD | Empty | 89.6 | 61.4 | 75.5 |
| | Empty + Solution-filled | 89.5 | 63.0 | 76.3 |
| Dense SSD | Empty | 95.0 | 67.5 | 81.2 |
| | **Empty + Solution-filled** | **99.5** | **90.8** | **95.2** |

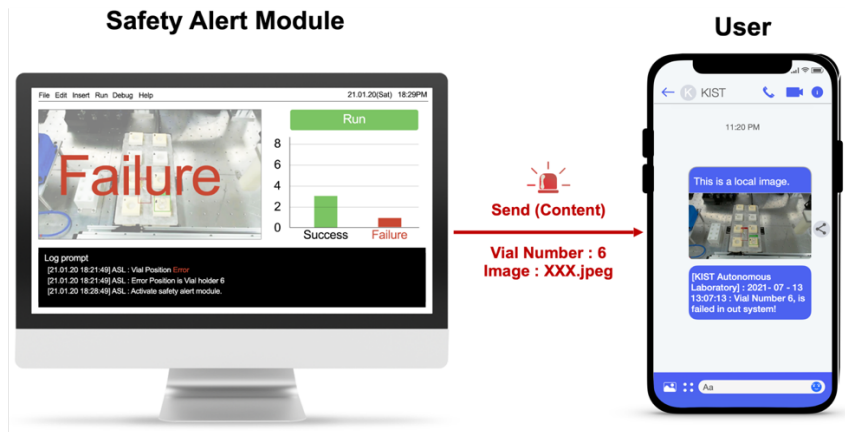

**Supplementary Figure S8. Scheme of the safety alert module.** The alert module remotely sends the scene image and related text such as event time and problematic vessel's information to the user's personal messenger platform.

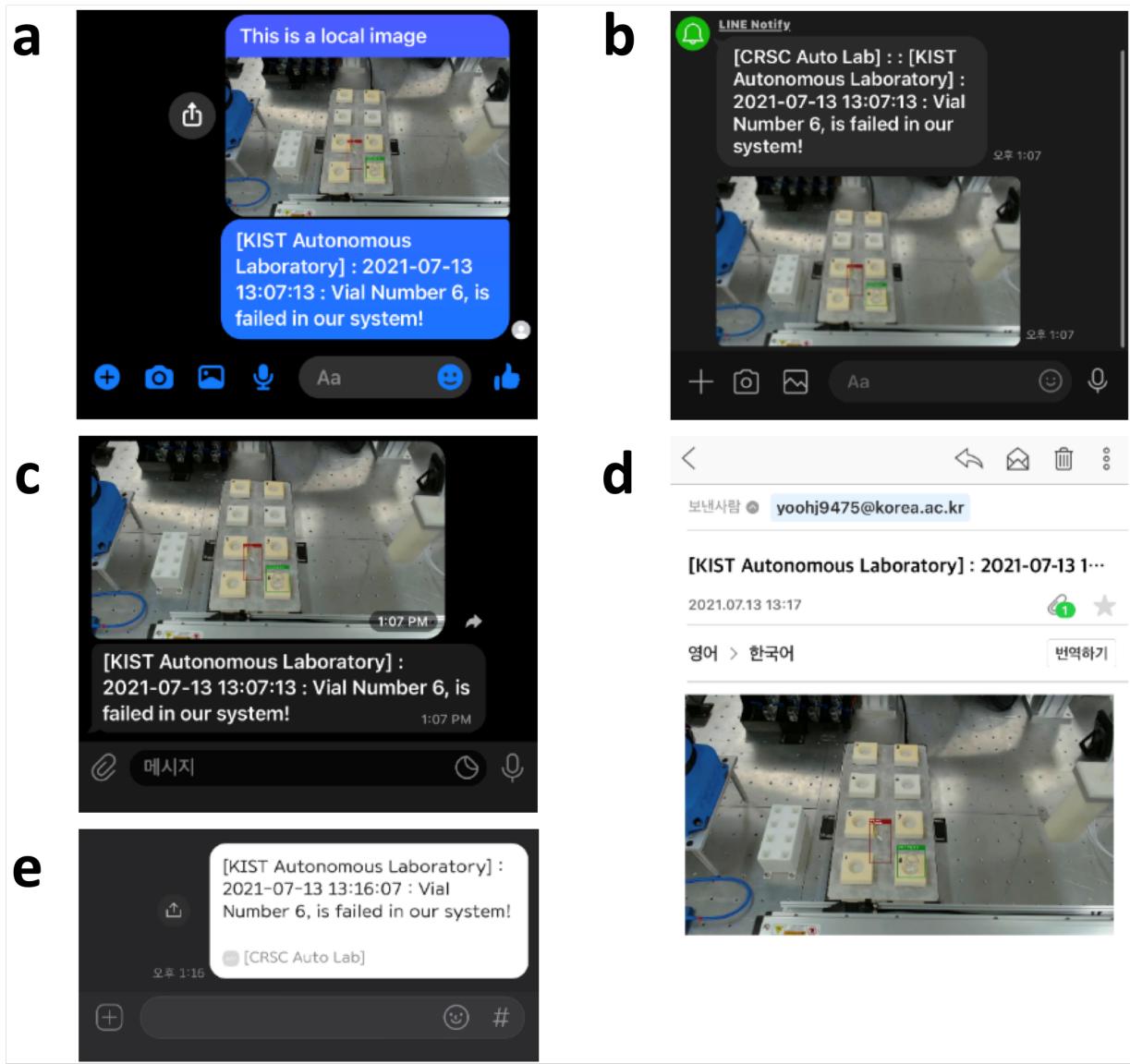

**Supplementary Figure S9. Demonstrations of the safety alert module for several messenger platforms.** (a) Facebook Messenger, (b) Line, (c) Telegram, (d) E-mail, (e) Kakaotalk. All the platforms' API are designed with Python open-source packages.